\pdfoutput=1
\documentclass{article}
\usepackage[english]{babel}
\usepackage[utf8]{inputenc}
\usepackage[letterpaper,top=2cm,bottom=2cm,left=3cm,right=3cm,marginparwidth=1.75cm]{geometry}
\usepackage{stmaryrd}
\usepackage{color,amsmath,amssymb, amsfonts, amstext,amsthm, latexsym}
\usepackage{booktabs}
\usepackage{multirow}
\usepackage{amsmath}
\usepackage{amsthm}
\usepackage{flushend, cuted}
\usepackage[title]{appendix}
\usepackage{bbm}
\usepackage{float}
\usepackage{epsfig, graphicx, graphics}
\usepackage{longtable}
\usepackage{cite}
\usepackage{subfigure}
\usepackage{caption}
\usepackage{bm}
\usepackage{authblk}
\usepackage[mathscr]{euscript}
\usepackage{diagbox}
\usepackage[ruled,vlined]{algorithm2e}
\usepackage{multirow}

\title{Early warning indicators via latent stochastic dynamical systems \vspace{-0.5em}}

\author[a,b,c]{Lingyu Feng }
\author[a,b,c]{Ting Gao \thanks{Corresponding author Ting Gao: tgao0716@hust.edu.cn; tinggao0716@gmail.com }}
\author[a,b,c]{Wang Xiao }
\author[c,d,e]{Jinqiao Duan \thanks{Jinqiao Duan: duan@gbu.edu.cn; duanjq@gmail.com } }

\affil[a]{School of Mathematics and Statistics, Huazhong University of Science and Technology, Wuhan 430074, China}
\affil[b]{Center for Mathematical Science, Huazhong University of Science and Technology, Wuhan 430074, China}
\affil[c]{Steklov-Wuhan Institute for Mathematical Exploration, Huazhong University of Science and Technology, Wuhan 430074, China}
\affil[d]{Department of Mathematics and Department of Physics, Great Bay University, Dongguan 523000, China } 
\affil[e]{Dongguan Key Laboratory for Data Science and Intelligent Medicine, Dongguan 523000, China}

\begin{document}

\maketitle

\begin{abstract}

Detecting early warning indicators for abrupt dynamical transitions in complex systems or high-dimensional observation data is essential in many real-world applications, such as brain diseases,  natural disasters,  and engineering reliability. To this end, we develop a novel approach: the directed anisotropic diffusion map that captures the latent evolutionary dynamics in the low-dimensional manifold. Then three effective warning signals (Onsager-Machlup Indicator, Sample Entropy Indicator, and Transition Probability Indicator) are derived through the latent coordinates and the latent stochastic dynamical systems. To validate our framework, we apply this methodology to authentic electroencephalogram (EEG) data. We find that our early warning indicators are capable of detecting the tipping point during state transition. This framework not only bridges the latent dynamics with real-world data but also shows the potential ability for automatic labeling on complex high-dimensional time series.  

\end{abstract}


\begin{quotation}
Leading Graph: In various practical scenarios, high-dimensional temporal processes are commonly observed. One particularly intriguing challenge involves the analysis of complex patterns in brain activities. The inherent complexity of modeling brain dynamics is exacerbated by the multiple EEG signals. To address this, we have introduced a directed anisotropic diffusion map method, which enables the extraction of latent dynamics in a lower-dimensional manifold. This approach proves valuable in identifying early warning signals for critical changes. Our framework establishes a connection between the latent transition dynamics modeled through stochastic dynamical systems and the real-world high-dimensional EEG data.
\end{quotation}

\section{Introduction}
\noindent 

High-dimensional evolutionary processes exist in many real-world systems, such as time-varying patterns of protein folding and unfolding, climate change, and artificial intelligence-assisted disease diagnosis \cite{mardt2018vampnets, yang2020tipping,mediano2022integrated}. Among these, one fascinating problem is studying complex brain activities over time. Due to the complexity of the human brain system, most of the multi-scale data-driven modeling approaches build surrogate models for brain activities. Electroencephalogram (EEG) signals are collected by dozens of electrodes placed on the scalp. Considering the low-dimensional representation from observations in electrophysiological experiments, neuroscientists have been investigating the latent dynamics of the brain via various dimension reduction techniques, leading to significantly enhanced performance by removing multi-collinearity and avoiding the curse of dimensionality.

One methodology for learning low-dimensional latent dynamics comes from the variational auto-encoder framework, which is able to simultaneously reduce dimension while learning latent dynamics\cite{feng2023learning}. However, the disadvantage is the weak explanation of effective coordinates of reduced systems.  {Therefore, other dimension reduction methods are highly investigated to represent the geometric structure of the dataset on a low-dimensional space \cite{berry2016local}.} Among the nonlinear methods,  {diffusion maps \cite{coifman2006diffusion}} are widely used to analyze dynamical systems \cite{nadler2006diffusion} and stochastic dynamical systems \cite{coifman2008diffusion} on low-dimensional latent space, determine order parameters for chain dynamics \cite{ferguson2010systematic}, classify person from electrocardiogram recordings \cite{sulam2017dynamical}, and so on. Also, dynamics of brain activity from task-dependent functional magnetic resonance imaging data are modeled by a three-step framework including diffusion maps in \cite{gallos2023data}.  
Furthermore, the directed anisotropic diffusion map  {\cite{coifman2005geometric}} improves upon the isotropic diffusion by incorporating the local information. It considers the original data structure and adapts the diffusion process accordingly via a directed term. To efficiently distinguish between epilepsy-related data and pre-ictal data in the latent space, we adopt the data-driven directed diffusion to construct an anisotropic diffusion map.

After dimension reduction through diffusion map, the latent dynamics could be learned in either deterministic or stochastic ways. For example, Talmon et al. study the inference of latent intrinsic variables of dynamical systems from observed output signals \cite{talmon2015manifold}.  Also, for partially observed data, Ouala et al. introduce a framework based on the data-driven identification of an augmented state-space model by a neural network \cite{ouala2020learning}. On the other hand, since stochastic differential equations are usually used to model nonlinear systems that involve uncertainty or randomness \cite{Duan2015AnIT}, it is of great interest to learn latent stochastic differential equations from real high dimensional data. To gain deeper insights into the latent characteristics, researchers frequently investigate the stochastic dynamical systems, which depend on the geometry and distribution of the latent space. For instance, Evangelou et al. combine diffusion map and Kramers–Moyal to construct an effective stochastic differential equation for electric-field mediated colloidal crystallization \cite{evangelou2023learning}. Besides, in general cases when the noise is non-Gaussian, there are various methods to realize the stochastic dynamical systems evolving in the latent space over time \cite{Li2021ExtractingGL}, as well as the corresponding transition dynamics \cite{dai2020detecting, Gao2015DynamicalIF}.

In real-world applications for complex disease diagnosis, early warning detection has attracted a great deal of attention. For instance, epilepsy is a neurological disease characterized by recurrent and unpredictable seizures, which are sudden bursts of abnormal brain activity. According to the World Health Organization, epilepsy affects around 50 million people worldwide. Early detection and intervention of epileptic seizures can help patients receive timely and appropriate treatment, which might prevent or minimize the negative impacts of seizures on their health and quality of life. Liu et al. present a model-free computational method to detect critical transitions with strong noise by dynamical network markers \cite{liu2015identifying}. Bury et al. develop a deep learning algorithm that provides early warning signals in systems by exploiting information on dynamics near tipping points \cite{bury2021deep}. 
In reality, the diagnosis of epilepsy requires experienced medical professionals. Analyzing extensive EEG data to determine the onset of an epileptic seizure is a time-consuming and labor-intensive process. Additionally, relying on visual recognition can result in doctors inaccurately assessing the timing of the epilepsy occurrence.
Therefore, in this paper, we propose to automatically detect the early warning signals from the latent stochastic dynamical system by introducing the Onsager-Machlup indicator, sample entropy indicator, and transition probability indicator.

In summary, the goal of the present study is to model and analyze the brain activity data from epileptic patients to identify early warnings of epileptic seizures automatically, with stochastic dynamical systems tools. Our main contributions are
\begin{itemize}
    \item Find the appropriate low-dimensional coordinates from time evolutionary high-dimensional data; 
    \item Learn the latent stochastic differential equation and dynamics on low-dimensional manifold;
    \item Explore effective early warning indicators for transitions in brain activity.
\end{itemize}

The remainder of this paper is structured as follows. In Section \ref{METHOD}, to obtain a better low dimensional representation, we adopt the directed anisotropic diffusion map with  { a data-driven directed term}. Following that, we extract the latent stochastic dynamical systems from the high dimensional dataset, and establish specific early warning indicators. In Section \ref{ES}, we validate our framework by numerical experiments on real EEG data. Finally, we summarize in Section \ref{CO}.


\section{Methodology}\label{METHOD}
\noindent

\begin{figure}[htp] 
    \centering
    \includegraphics[width=10cm]{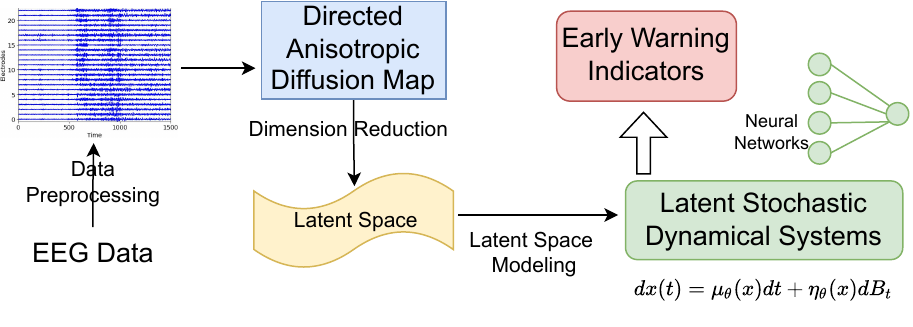}
    \caption{A schematic diagram of our framework. }
    \label{nn}
\end{figure}

In this section, we introduce the main procedures in our workflow of exploring early warning signals from latent dynamical systems  {(Fig. \ref{nn})}. 

\subsection{Anisotropic Diffusion Map}
\noindent

The core idea of diffusion map is a time-dependent diffusion process, i.e., a random walk on the dataset where each hop has the associated transition probability. When the diffusion process runs, for each step, we can calculate the distance over the underlying geometric structure. To evaluate the connectivity strength between two data points or the probability of jumping from $x$ to $y$ in one step of the random walk, we need to introduce a diffusion kernel.

Let $X$ be the dataset in $\mathbf{R}^n$, and define a kernel $k(x,y)$ on $X \times X$.  {Note that $k(x,y)=1$ if and only if $x=y$, and $k(x,y)$ has exponential decay with distance}. This kernel captures some geometric features from the data and the probability of transitions between $x$ and $y$. The technique to obtain the latent diffusion space is known as the normalized graph Laplacian construction\cite{chung1997spectral}. Assume that a set of points on or near a $d$-dimensional manifold are embedded in a high-dimensional space $\mathbf{R}^n$, and the manifold exists. The map $\Phi: \mathcal{M} \rightarrow \mathbf{R}^d$ learns the geometric structure of the data $X$ sampled from $\mathcal{M}$ with the density $q$.

The isotropic diffusion computes all directions equally. It means that all variables in dataset play the same role. However, some variables might be more important than others under the prior knowledge of the dataset. The anisotropic diffusion is sometimes more useful to locally separate the variables and provides early warnings.

To design a diffusion process adapted to the dataset, consider the directed kernel,   {mentioned in \cite{coifman2005geometric,lafon2004diffusion},}
\begin{equation}\label{dker}
      {k_\varepsilon(x,y) = \exp(-\frac{\parallel x-y\parallel^2}{\varepsilon}-\frac{\parallel \langle\nabla_x f,x-y\rangle\parallel^2}{\varepsilon^2}),}
\end{equation}
where $x,y \in \mathbf{R}^n$,   {$\nabla_x f$ is the gradient of $f$ (defined on dateset $X$) at the point $x$}, and $\parallel\cdot\parallel$ denotes the Euclidean norm.  {Here, $f$ could be determined by the intrinsic geometric and dynamical characteristics of the data (see more details in Section \ref{Exdimen} Eq.(\ref{empiricalFun})).}

 {In the later experiment, we will show the benefit of anisotropic directed diffusion map, since for real data applications, it is more capable of capturing the intrinsic distribution shift of the data, as a return, generating more effective early warning indicators from the low-dimensional latent dynamics.}

The weight matrix $K$ corresponds to an anisotropic kernel on the observable space $X$.
By integrating over the second variable $k_{\varepsilon}$, the density $d_{\varepsilon}$ is represented by
\begin{equation*}
    d_{\varepsilon}(x) =  \int_{X} k_{\varepsilon}(x,y)q(y)dy,
\end{equation*}
and define 
\begin{equation}
    p_{\varepsilon}(x,y) =  \frac{k_{\varepsilon}(x,y)}{d_{\varepsilon}(x)}.
\end{equation}
The directed operator $P_{\varepsilon}$ is defined by
\begin{equation}
    P_{\varepsilon}f(x)=\int_X p_{\varepsilon}(x,y)f(y)q(y)dy.
\end{equation}

Dealing with finite data points, we should approximate integrals by finite sums and can describe the latent reduced order space by the spectral analysis. 
For most applications, we consider the bounded part will consist of a finite number of points. Here, the eigenvalues of the directed diffusion operator are distributed between $0$ and $1$. In other words, we have $1\ge\lambda_1\ge\lambda_2\ge\cdots$. The first $d$ eigenvalues and the corresponding eigenvectors capture the main features. The directed anisotropic diffusion map is given by $\Phi(x) = [\lambda_1\phi_1(x)\dots\lambda_d\phi_d(x)]^T$, where $\{\phi_i\}_{i=1}^d$ are eigenvectors. Several benefits are obtained by dimension reduction. It is well known that high dimensionality often degrades the dynamical performance, necessitating the use of lower-dimensional representations. Additionally, in the latent space, there is a significant reduction in the time required to train a model.

\subsection{Extracting Latent Stochastic Dynamical System}\label{loghood}
\noindent 

In order to depict the broader spectrum of nonlinear dynamics in the latent space, stochastic differential equations are preferred to carry the low dimensional latent dynamics of the high-dimensional data. The drift term (vector field) characterizes the deterministic aspect of the system. While the diffusion term (fluctuation) pertains to the stochastic component of the system under noise such as Brownian motion or L\'evy process.

Let $z(t)$ be a stochastic  {process} governed by the stochastic differential equation
\begin{equation}\label{sdeq}
    dz(t) = \mu(z(t))dt +\eta(z(t))dB_t,
\end{equation}
where $\mu: \mathbf{R}^d \rightarrow \mathbf{R}^d$ is the drift term, $\eta: \mathbf{R}^d \rightarrow \mathbf{R}^{d\times d}$ is the diagonal diffusivity matrix, and $B_t$ is $\mathbf{R}^d$ independent Brownian motion. Inspired by deep learning architecture\cite{dietrich2023learning}, we construct two neural networks $\mu_\theta$ and $\eta_\theta$ to estimate $\mu$ and $\eta$, whose inputs and weights are represented by the basis functions and coefficients, respectively. 

In fact, we have access to a set of lower-dimensional points $\{z(t_i)\} \in \mathbf{R}^d$ by directed anisotropic diffusion map. Then the set of snapshots $\{z(t_{i+1}),z(t_i),\delta_t\}$, $\delta_t = t_{i+1}-t_{i}$ is available. Over the small time interval $[t_{i},t_{i+1}]$, the stochastic differential equation has a discrete form
\begin{equation*}
    z(t_{i+1}) = z(t_i)+\mu(z(t_i))\delta_t +\eta(z(t_i))\delta_{B_t},
\end{equation*}
where $\delta_{B^j_t} = B_{t_{i+1}}-B_{t_{i}} \sim \mathcal{N}(0,\delta_t)$, $j=1,\ldots,d$. Conditioned on $(z(t_i),\delta_t)$, the point $z(t_{i+1})$ obeys a multivariate normal distribution $\mathcal{N}(z(t_i)+\delta_t\mu(z(t_i)),\delta_t\eta^2(z(t_i))).$
Hence, the neural networks are trained by minimizing the loss function $\mathcal{L}(\theta)$, which is defined as
\begin{equation}
    \mathcal{L}(\theta) = \frac{(z(t_{i+1})-z(t_{i})-\delta_t\mu_\theta(z(t_i)))^2}{\delta_t\eta^2_\theta(z(t_i))}+\log \mid \delta_t\eta^2_\theta(z(t_i)) \mid+\log(2\pi).
\end{equation}
Moreover, depending on the tail property of the real dataset, the latent space modelling can be extended to stochastic differential equations driven by non-Gaussian noise  \cite{fang2022end}.

\subsection{Early Warning Indicators}
\noindent

 {After obtaining the latent stochastic dynamical system from high dimensional data, we can now identify some abrupt critical transitions, through indicators created by Onsager-Machlup action functional, sample entropy, and transition probability.} 

\subsubsection{Onsager-Machlup Indicator}
\noindent

Consider the following stochastic differential equation in the state space $\mathbb{R}^d$  $$dz_t=\mu(z_t)dt+\eta dB_t,\quad t>0,$$ 
where $\mu:\mathbb{R}^d\to\mathbb{R}^d$, $B_t$ is a standard $d$-dimensional Brownian motion, and $\eta$ is a diagonal matrix with positive constants. 
Onsager and Machlup are the first to consider the probability of paths of a diffusion process as the starting point of a theory of fluctuations \cite{onsager1953fluctuations}. The outcome of a functional integral over the process paths is subsequently termed the Onsager-Machlup function, denoted by
$$S_X^{OM}(\psi)=\frac12\int_0^T\left[\frac{|\dot{\psi}(s)-\mu(\psi(s))|^2}{\eta^2}+\nabla \cdot \mu(\psi(s))\right]ds,$$
where $\psi$ is a curve in $\mathbb{R}^d$.

Let time series $z(t)$ be a stochastic variable governed by the stochastic differential equation in the latent space. Recalling Sec. \ref{loghood}, two neural networks $\mu_\theta$ and $\eta_\theta$ are constructed to estimate $\mu$ and $\eta$, whose inputs and weights are represented by the basis functions and coefficients, respectively.
Here, we define the Onsager-Machlup  {functional} by
\begin{equation}\label{OMindi}
    {OM}(l,t)=\frac12\int_{t-l+1}^t\left[\frac{|\dot{z}(s)-\mu_\theta(z(s))|^2}{\eta_\theta^2}+\nabla \cdot \mu_\theta(z(s))\right]ds,
\end{equation}
where $\dot{z}$ is the derivative function written as  {$\dot{z}(s) \approx \frac{z(s+ds)-z(s-ds)}{2ds}$ with the time step $ds$.}

\subsubsection{Sample Entropy Indicator}
\noindent

Sample entropy \cite{richman2000physiological}, serves as a statistical metric for quantifying the complexity of time series data. It is particularly effective in capturing self-similarity and irregularity, making it a robust tool for analyzing complex systems. For $d$ interdependent time series, we introduce the multidimensional sample entropy to measure the latent system complexity \cite{meng2020complexity}. Specifically, it approximately equals to the negative natural logarithm of the conditional probability that two sub-sequences similar for consecutive data points remain similar for the next $p$ time points.

For $d$ time series $z^\alpha(t) \ ( \alpha=1,\ldots,d)$, the sequence of $l$ time to compute the sample entropy is expressed by $Z(t_l)=\{z^\alpha[t_1:t_l],\ \alpha=1,\ldots,d\}$, where $z^\alpha[t_1:t_l] = \{z(t_1),\ldots, z(t_l)\}$. There forms the vectors $Z^\alpha_k(m,q) =\{z^\alpha[kq+1:kq+m]\}\subset Z(t_l) \ (k=1,2,\ldots).$ The distance between two vectors is defined to be the Chebyshev distance $$d(Z^\alpha_i(m,q),Z^\beta_j(m,q))=\max\{|x^\alpha(iq+s)-x^\beta(jq+s)|, 1\leq s\leq m\}.$$ If the superscripts $\alpha=\beta$, the subscripts $i\neq j.$
Assume that two vectors are close if their distance $d(Z^\alpha_i(m,q),Z^\beta_j(m,q))<r\max\{\sigma^\alpha,\sigma^\beta\}$, where $r$ is the tolerance for accepting matches and $\sigma^\alpha$ and $\sigma^\beta$ are the standard deviations, allowing measurements on dataset with different amplitudes to be compared. Now we consider a set $A(m+p,q)$ of all vectors satisfying $Z^\alpha_k(m+p,q)\subset z^\alpha[t_1:t_l]$ and $\alpha = 1,\ldots,d$. Note that the maximum of $k$ in $A(m+p,q)$ is $n_z$. To make the number of vectors of length $m$ equal to that of length $m+p$, the set $B(m,q)$ is characterized as $\{Z^\alpha_k(m,q): Z^\alpha_k(m,q) \subset z^\alpha[t_1:t_l], 1\leq \alpha \leq d, 1\leq k\leq n_z\}$.
Define $B(m,q,r,l)$ as the number of close vectors in $B(m,q)$. Similarly, denote the number of close vectors in $A(m+p,q)$ by $A(m+p,q,r,l)$. Hence, the sample entropy $SampEn(m,p,q,r,l)$ for multidimensional system is represented by $-\log\left(\frac{A(m+p,q,r,l)}{B(m,q,r,l)}\right)$
in the time interval $[t_1,t_l]$.
The parameters $l,\:m,\:p,\:q$ and $r$ must be fixed for each calculation. So in time window $[t-l+1,t]$, we have the entropy indicator
\begin{equation}\label{SEeq}
    SampEn(t) = -\log \left(\frac{A(m+p,q,r,l)}{B(m,q,r,l)}\right).
\end{equation}

\subsubsection{Transition Probability Indicator}
\noindent

The transition probability between two states is defined in \cite{Bolhuis2002TransitionPS}. The phase space for the latent variable is separated into two regions $A$ and $B$ and the ratio $\mbox{TP}_{AB}$ is defined as
\begin{equation}\label{eqtrp}
    \mbox{TP}_{AB}(t) = \frac{\langle I_A(z(0))I_B(z(t))\rangle}{\langle I_A(z(0))\rangle},
\end{equation}
where $I_A$ is an indicator function satisfying $I_A(z)= 1$ if $z \in A$, and $I_A(z) = 0$ if $z \notin A$ and $\langle\cdot\rangle$ denotes the ensemble average. The ratio $\mbox{TP}_{AB}$ represents the probability of finding the system in region $B$ after time $t$ under the condition $z(0) \in A$. In the following, if the probability is more than $0.5$, we will give an early warning due to higher transition probability.

\section{Experiments}\label{ES}
\noindent 

We now apply our method to the real datasets collected from epileptic patients. Our objective is to analyze the latent dynamics and illustrate our early warning indicators.

\subsection{Data Preprocessing}
\noindent

\begin{figure}[htp] 
    \centering
    \includegraphics[width=4cm]{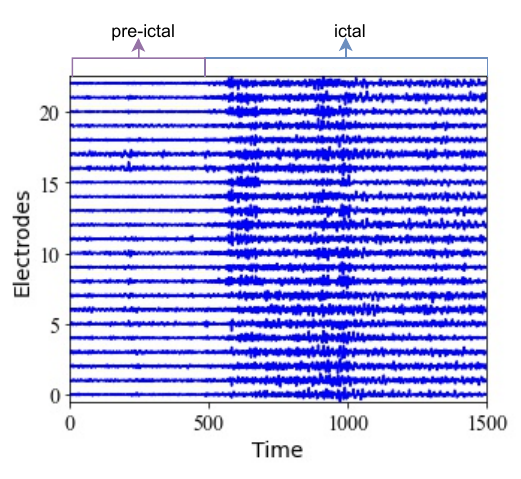}
    \caption{An example: the EEG data of a patient. }
    \label{eeg}
\end{figure}

Electroencephalogram (EEG) captures signals from electrodes, thereby recording the electrical activity during epileptic seizures.  The EEG dataset encompasses both pre-ictal and ictal data, organized in a matrix format of channels over time, with a sampling rate of 256 points per second. Here, each electrode signal undergoes rescaling to an interval of $[-0.5, 0.5]$ and is subsequently subsampled by averaging every sixteen points, resulting in the dataset  {with the time step $\Delta t = 0.0625$}. In Fig. \ref{eeg}, there is a sample image in which the doctor labels $Time = 500$ as the separation time for the pre-ictal and ictal periods.

\subsection{Dimension Reduction with Directed Anisotropic Diffusion Map}
\label{Exdimen}
\noindent

To construct the diffusion matrix, the data from all electrodes at each time point is viewed as a high-dimensional node in the diffusive graph. Notably, electrode signals display abnormal fluctuations preceding the onset of an epileptic seizure. Timely detection of abnormalities in electrical signals holds the potential to provide early warnings for epileptic seizures. We suggest identifying early warnings within a lower-dimensional space through the application of directed anisotropic diffusion maps.

The diffusion matrix is constructed by the diffusion kernel and shows the transition among all high-dimensional nodes in the diffusive graph, which exhibits the transition probability from one point to another.   
 {To find the appropriate $\nabla_x f$ in a data driven way, we introduce the gradient flow in $\mathbf{R}^n$,}
\begin{equation}
\label{empiricalFun}
 {\dot{x}=-\nabla_x f(x)}.
\end{equation}
 {Here, the gradient flow $\nabla_x f$ at $x_i = x_{t_{i}}$ is approximated by $\nabla_x f(x_{t_{i}})\approx -\frac{(x_{t_{i+1}}-x_{t_{i-1}})}{2\Delta t}$, which will be substituted into Eq.(\ref{dker}) for kernel calulation.}
Note the stochastic derivative is approximated similar to the finite difference formula, indicating the dynamic trend of stochastic process $x(t)$. 
\begin{figure}[htp] 
    \centering
    \includegraphics[width=12cm]{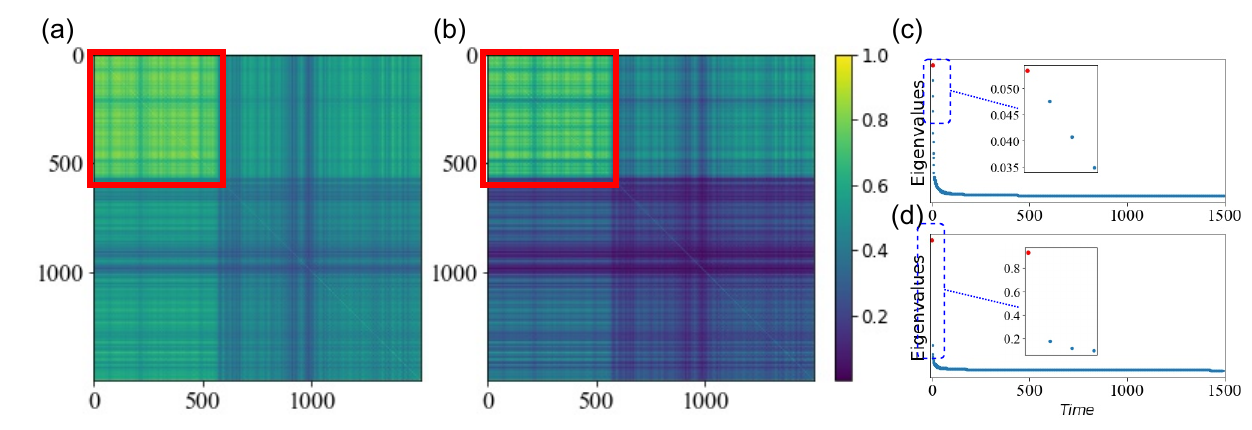}
    \caption{Comparison of diffusion map and directed anisotropic diffusion map.  {The first two columns (a) (b) represent the kernel matrices, and (c) (d) give the eigenvalues for diffusion map and directed anisotropic diffusion map respectively.} For all the cases, we set $\epsilon = 1$. Note that (a) (c) are constructed  {without the directed term}, while (b) (d) are calculated  {with the data driven directed term learned by Eq.(\ref{empiricalFun}) }.}
    \label{tmeig}
\end{figure}

Then we  {can obtain the kernel matrices from the preprocessed dataset}. As shown in Fig. \ref{tmeig}, most points have a probability greater than $0.5$ with another point in $(a)$, indicating that the transition probability to other points is relatively equivalent. In $(b)$, the data points in the red frame have higher transition probabilities to each other, while the transition probabilities to the points outside gradually decrease.  {Through directed diffusion, the transition probability of sample points from inside the red box to the outside is reduced, and similar when points transit from outside to the inside. In this way, the diffusion intensity between points inside and outside the red box is forced to attenuate, so the two regions ('pre-ictal' and 'ictal' stages) could be better separated.}

The diffusion matrix has a discrete sequence of eigenvalues $\{\lambda_i\}_{i\ge 1}$ and eigenvectors $\{\phi_i\}_{i\ge 1}$ such that $1 \ge \lambda_1\ge\lambda_2\ge\cdots$. 
Note that for isotropic diffusion, the maximum eigenvalue is equal to 1, while for directed anisotropic diffusion, the largest eigenvalue is less than 1. Here, to illustrate the whole framework of our method in a straightforward way, we choose the spectral gap to help us find the effective dimension of latent space. In Fig. \ref{tmeig}, (c) shows the first four eigenvalues of the diffusion map have similar gaps, while the first two eigenvalues of the directed anisotropic diffusion map in (d) have a bigger spectral gap than others. Thus we choose the latent space to be one-dimensional under directed anisotropic diffusion map. Considering some of the large eigenvalues are associated with the harmonics of earlier ones, there are more algorithms to find the effective embedding dimension \cite{dsilva2018parsimonious}, which can be further studied in the future.

\begin{figure}[htp] 
    \centering
    \includegraphics[width=8cm]{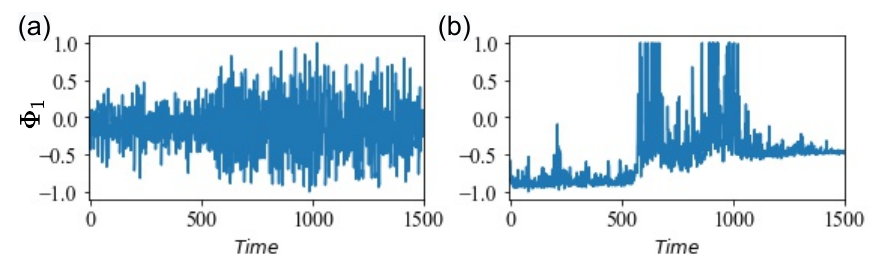}
    \caption{Comparison of low-dimensional embedding patterns with (a) diffusion map and (b) directed anisotropic diffusion map.}
    \label{cod}
\end{figure}

 {Now we compare the data projected onto the latent coordinate $\Phi_1(x) = \lambda_1\phi_1(x)$ with two different diffusion maps. In Fig. \ref{cod}, both the projected data shows distribution shift at $Time \approx 600$. However, in $(b)$, the data after the 'shift' has bigger variance and large deviated mean value compared with that before $Time \approx 600$. While in $(a)$, the data points before and after the shift have almost the same expectation, with bigger variance in the latter time period. We can see these two different patterns affect the dynamics of latent stochastic differential equations in the next section. }

\subsection{Latent Stochastic Dynamical Systems}\label{EXlatent}
\noindent

In this section, we aim to elucidate the connection between the latent stochastic differential equation and the original high-dimensional time evolutionary data.

Following dimension reduction through diffusion map, we obtain the projected dataset onto the first diffusion coordinate $\Phi_1$. The projected data plotted over time displays the transition between two states: pre-ictal state and ictal state. Therefore, we assume that the stochastic differential equation in the latent space has two meta-stable states and rescale the data to an interval of $[-1,1]$ for convenience. Given data in Fig. \ref{cod} (b), we would like to obtain the latent stochastic differential equation in the form of Eq. (\ref{sdeq}). To get meaningful representations of drift and diffusion coefficients, we employ neural networks to approximate the latent equations. The inputs of networks are the polynomial basis functions of the coordinates and the time step is $0.0625$. The latent space contains $1500$ time points and $1499$ pairs of $\{(z(t_{i+1}),z(t_i))\}$ are available. We randomly select $1200$ pairs to train the networks and validate the identification accuracy on the rest $299$ pairs.

To find the explicit form of drift and diffusion coefficients in the unknown stochastic differential equations, all the unknown drift and diffusion terms are parameterized with neural networks. Then using the loglikelihood method in Sec. \ref{loghood}, the learnt stochastic differential equation under the coordinate $\Phi$ has the form
\begin{equation}\label{latentSDE} 
    dz(t) = \mu(z(t))dt + {\eta}(z(t))dB_t,
\end{equation}
where  {for diffusion map, $\mu(z) =  -0.0650+  0.4767z +0.5048z^2 -1.5105z^3$ and $\eta(z)=-0.4711$,} and for directed diffusion map,
 {$\mu(z) =  0.1849+  0.5058z -0.6381z^2 -1.4880z^3$ and $\eta(z)=-0.4896$}.

\begin{figure}[htp] 
    \centering
    \includegraphics[width=8cm]{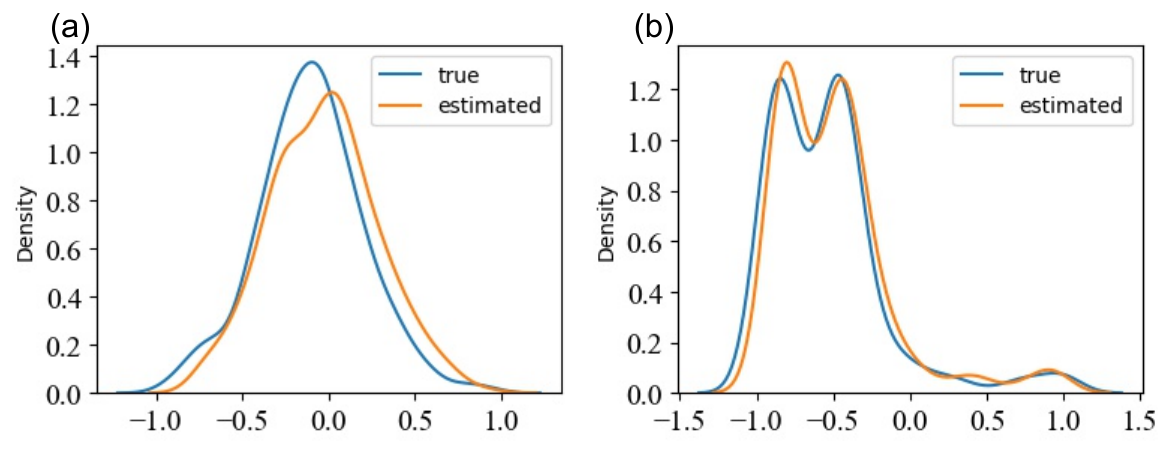}
    \caption{The validation of the neural network approximation with regard to $\Phi_1$.  {(a) Diffusion map.} (b) Directed anisotropic diffusion map.}
    \label{vald}
\end{figure}

Moreover, we adopt density as the validation metric for the testing data (consisting of $299$ pairs). In Fig. \ref{vald}, we compare the density of the estimated $z(t_{i+1})$ with the ground truth from corresponding projected data points in Fig.\ref{cod}. The alignment between our approximation and the true data density is evident. Additionally, the presence of two peaks in the density in Fig. \ref{vald} (b) signifies the existence of two meta-stable states in the projected coordinate by directed anisotropic diffusion map.  {However, both the densities in (a) only have the single peak. }

\subsection{Early Warning Indicators}
\noindent
 
In this section, we present the time varying indicators resulted from transitions in the latent space.\\

\textbf{Onsager-Machlup Indicator and Sample Entropy Indicator}
\noindent

\begin{figure}[htp] 
    \centering
    \includegraphics[width=15cm]{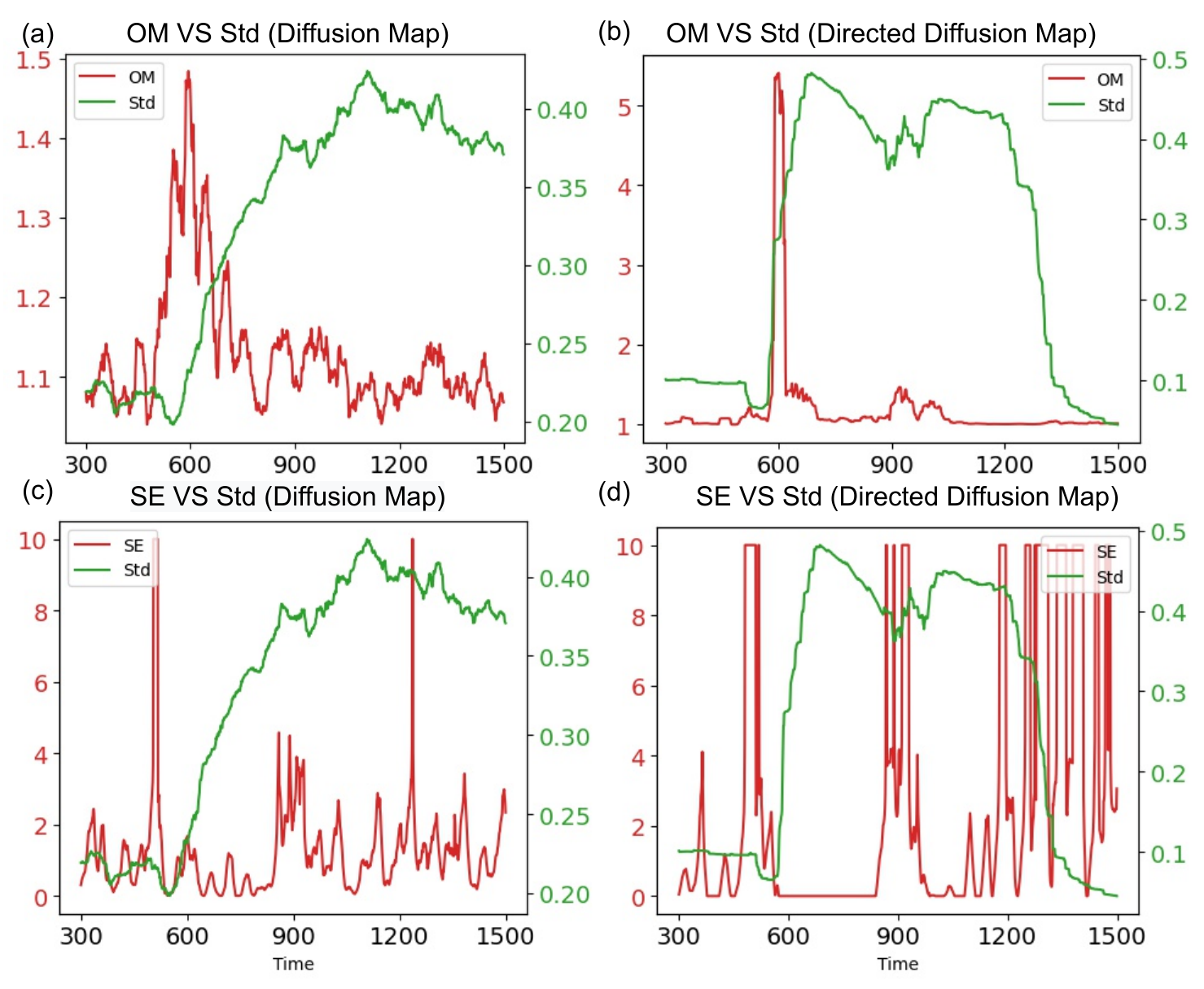}
    \caption{The Onsager-Machlup indicator (OM)  and the sample entropy (SE) of raw data projected onto coordinates $\Phi_1$ by   {diffusion map} and the directed anisotropic diffusion map, compared with the standard deviation(Std)(green).  {(a) OM(Diffusion map). (b) OM(Directed diffusion map). (c) SE(Diffusion map). (d) SE(Directed diffusion map).}}
    \label{omse}
\end{figure}

For $\Phi_1$, we employ a time series comprising $1500$ data points to calculate the indicator over time and compare it with the standard deviation.  {The time window $l$ is set to be $300$}, moving forward in each time unit $\Delta t =0.0625$.  {We propose the Onsager-Machlup indicator as the ratio of two Onsager-Machlup functionals (Eq. (\ref{OMindi})) in each time window, equal to $\frac{OM(l,t)}{OM(0.9l,t-0.9l)}$.} When applying the Eq. (\ref{SEeq}) to compute the sample entropy, the parameters are kept constant at $m=5$, $p=10$, $q=1$, and $r=2$. The values of the two indicators over time are red in Fig. \ref{omse} and the standard deviation is green.

 {First, considering the two patterns of projected data in Fig. \ref{cod}, we can see that the standard deviations (green) in Fig. \ref{omse} with diffusion map (left column) and directed diffusion map (right column) have interpretable shapes. One interesting observation we would like to mention is that the directed diffusion map also helps the standard deviation(Std) indicator sharply change at $Time \approx 600$, which is quite different from the slowly increasing shape of Std under the diffusion map in the left column. }

 {Next, in the top row of Fig. \ref{omse}, both the Onsager-Machlup indicators with two kinds of diffusion maps show early warning signals before the standard deviation indicator booms up. The difference is that the directed diffusion map (b) makes the signal more narrow and peaky, while the diffusion map (a) makes the signal noisy and slowly changing at $Time \approx 600$. Furthermore, in the bottom of Fig. \ref{omse}, both the sample entropy indicators with two types of diffusion maps present earlier warnings,  compared with standard deviation. We can even find the benefit from directed diffusion map (d) that two more peaks of the entropy indicator ($Time \approx 900$ and $Time \approx 1200$) align well with the drop of the standard deviation.} Consequently, the detection of change points can assist doctors in distinguishing pre-ictal and ictal periods, facilitating automated labeling processes. \\

\textbf{Transition Probability Indicator}
\noindent

\begin{figure}[htp] 
    \centering
    \includegraphics[width=8cm]{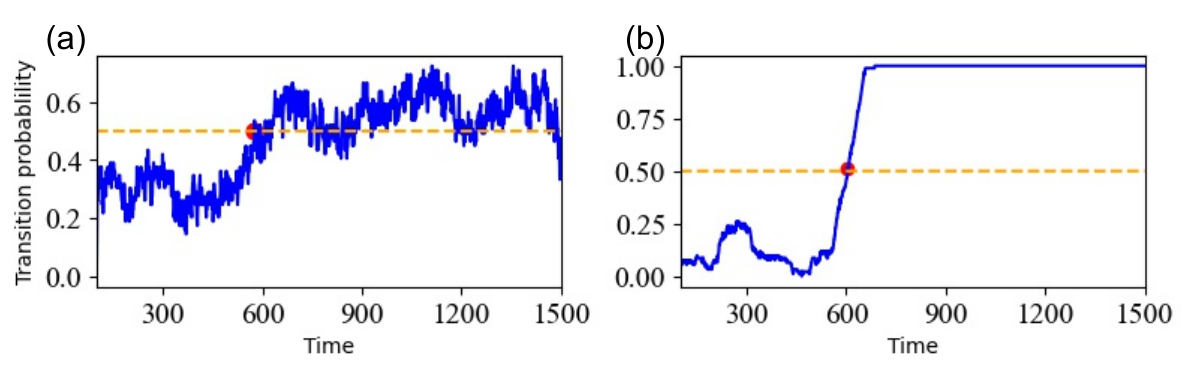}
    \caption{The transition probability of raw data projected onto coordinate $\Phi_1$. The red dot indicates the transition probability equal to $0.5$.  {(a) Diffusion map.} (b) Directed anisotropic diffusion map.}
    \label{tr2}
\end{figure}

Fig.\ref{tr2} shows the transition probability, assigning the width of each time window as $100$. The first $100$ points are selected as starting points, generating $100$ samples. For the directed anisotropic diffusion map, the split point is set as $-0.75$, delineating regions $A = (-\infty, -0.75]$ and $B = (-0.75, \infty)$. Similarly, the regions are $A = [-0.25, 0.25]$ and $B = \mathbf{R}\backslash A$ for the diffusion map. In Fig. \ref{tr2}, we calculate the transition probability by Eq. (\ref{eqtrp}) of state for each time step $t$ starting from the initial state. Assuming a transition occurs when the probability surpasses $0.5$, we designate the point with a probability equal to $0.5$ as the warning point.  {The warning points appear at $Time \approx 600$ for both two types of diffusion maps. However, the transition probability of the directed anisotropic diffusion map is much smoother, representing the two regions are better separated more robustly.}

\subsection{ {Validation}}
\noindent

 We select the epileptic data of the same patient at a different visiting time to the hospital to serve as the validation set for our framework. To conduct out-of-sample verification for the indicators, we project new data points onto the embedded manifold  {by the geometric harmonics method \cite{coifman2006diffusion,coifman2006geometric}}. Specifically, the coordinates derived from the directed diffusion maps serve as a basis for the embedded manifold, and the formulation of the latent extension for the new point $x^{*}$ is carried out by $\Phi(x^{*})=[\Phi_1(x^{*}) \ldots \Phi_d(x^{*})]$, where
\begin{equation*}
     {\Phi_i(x^{*}) = \frac{1}{\lambda_i}\Sigma_{j}\frac{k_\varepsilon(x^{*},x_j)}{\Sigma_{j}k_\varepsilon(x^{*},x_j)}\Phi_i(x_j), \quad i=1,\ldots,d.}
\end{equation*}

\begin{figure}[htp] 
    \centering
    \includegraphics[width=12cm]{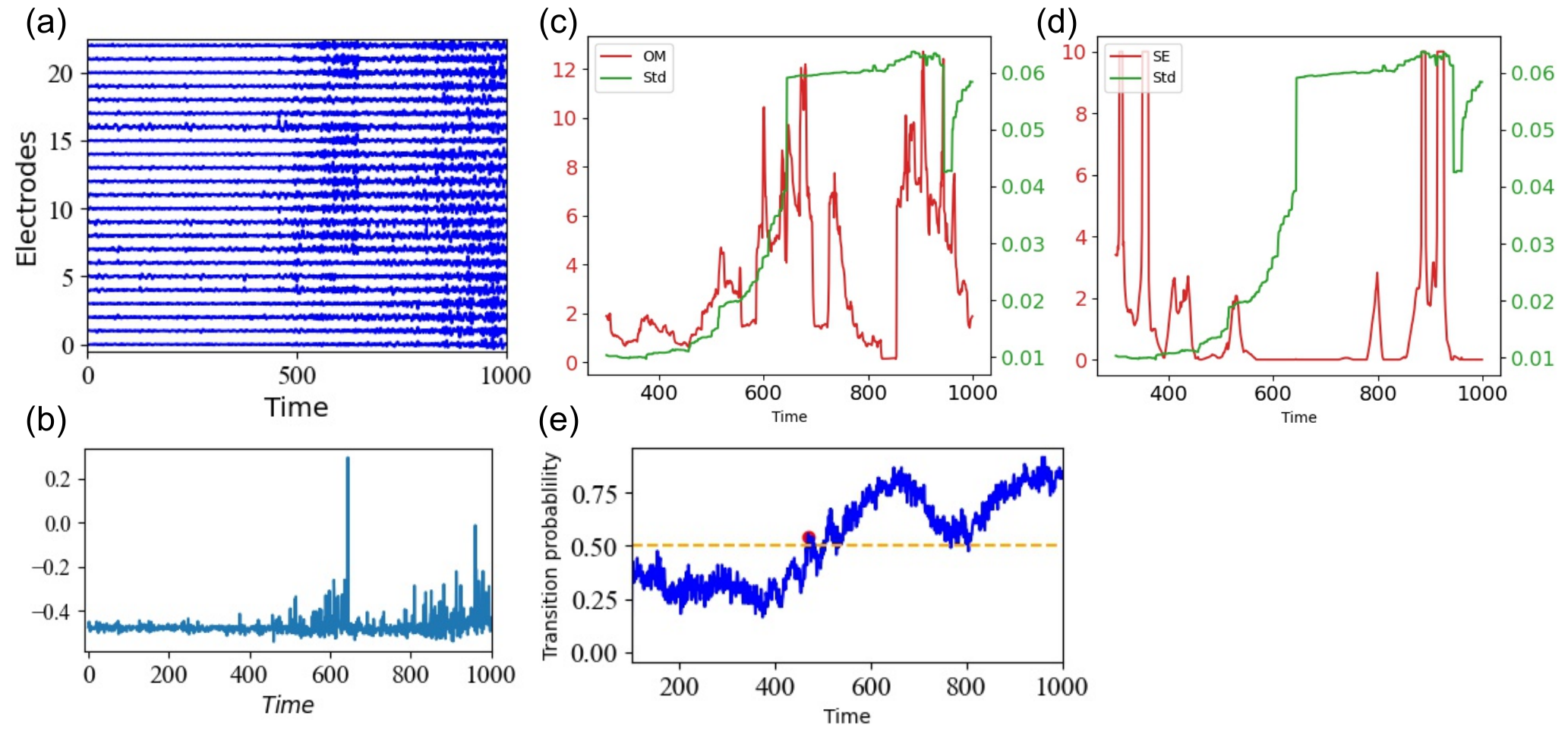}
    \caption{ {The out-of-sample points validate the effectiveness of our framework. (a) Validation dataset. (b) Embedding pattern. (c) OM indicator(red) compared with the Std(green). (d) SE indicator(red) compared with the Std(green). (e) The transition probability.}}
    \label{newp}
\end{figure}

 {The validation set undergoes the same data preprocessing steps, and the new dataset (Fig. \ref{newp} (a)) comprises $1000$ data points, each with a time step of $0.0625$. The initial $500$ points represent pre-ictal data, followed by ictal data, labeled by medical professionals. The low-dimensional embedding of the new sample, projected onto the latent space, is shown in Fig. \ref{newp} (b). 
Moreover, three indicators are presented in (c), (d) and (e), with the same time window length $300$ and the latent stochastic differential equation learned in Section \ref{EXlatent} under directed diffusion map. Notably, the upward trend of the Onsager-Machlup indicator outperforms that of the standard deviation in (c), and the sample entropy exhibits an abnormal spike before $Time = 400$ in (d). Unfortunately, the transition probability indicator in (e) exhibits suboptimal performance. 
Despite the imperfect performance of one indicator, the new validation set underscores the feasibility of employing our method for the automated detection of abnormal signals in epileptic seizures, circumventing the need for manual diagnosis by medical professionals.}

\section{Conclusion}\label{CO}
\noindent

In this paper, we propose an efficient method for automatically identifying abnormalities in EEG signals. Our approach leverages directed anisotropic diffusion map for dimensionality reduction and establishes stochastic dynamical system in the low-dimensional latent space,  { which facilitates us to design early warning signals more efficiently. Besides, we also apply the framework to new dataset as validation.}

Directed diffusion is employed to differentiate ictal data from pre-ictal data under the notion that information can flow in a specific direction. This technique serves as an extension of the diffusion map algorithm, specifically tailored to extract and represent the underlying structure and relationships in high-dimensional data. A crucial factor in defining a reduced latent space is the presence of a spectral gap among the eigenvalues. In this paper, we focus on effective coordinates that capture the dynamic characteristics of the original high-dimensional data.

Early warning of epileptic seizures is of paramount importance for epileptic patients. The abrupt change is caught for early warning in the latent space, where normal state and ictal state can be viewed as two meta-stable states. The ability to identify transitions between meta-stable states plays a pivotal role in predicting and controlling brain behavior. We derive three effective warning signals—namely, the Onsager-Machlup indicator, the sample entropy indicator, and the transition probability indicator—utilizing information from the latent coordinates and the latent stochastic dynamical systems. These indicators enhance the robustness and accuracy of early warning systems for epileptic seizures.  {Furthermore, the computational cost of calculating these indicators from low dimensional data is much lower than that of the original high-dimensional data.} This framework of learning latent stochastic systems and detecting abnormal dynamics has the potential to extend to general scenarios for other complex high dimensional time evolutionary data.

\section*{Acknowledgements}
We would like to thank Professor Quanying Liu and her students Yinuo Zhang, Zhichao Liang for providing the data and helpful discussion. This work was supported by the National Key Research and Development Program of China (No. 2021ZD0201300), the National Natural Science Foundation of China (No. 12141107), the Fundamental Research Funds for the Central Universities (5003011053), the Fundamental Research Funds for the Central Universities, HUST: 2022JYCXJJ058 and Dongguan Key Laboratory for Data Science and Intelligent Medicine.

\section*{Data availability}
The data that support the findings of this study are openly available in GitHub at 
https://github.com/LY-Feng/EW-in-latent-SD.

\bibliographystyle{unsrt}
\bibliography{ref}

\end{document}